%% file: CSNDSP_2026_ZVLDDD.tex
\documentclass[conference]{IEEEtran}
\IEEEoverridecommandlockouts

\usepackage{cite}
\usepackage{amsmath,amssymb,amsfonts}
\usepackage{algorithmic}
\usepackage{graphicx}
\usepackage{textcomp}
\usepackage{xcolor}
\usepackage{tabularx}
\usepackage{booktabs}
\usepackage{multirow}
\usepackage{colortbl} 
\usepackage[caption=false,font=footnotesize]{subfig}

\def\BibTeX{{\rm B\kern-.05em{\sc i\kern-.025em b}\kern-.08em
    T\kern-.1667em\lower.7ex\hbox{E}\kern-.125emX}}

\newcommand{\e}{\mathbf{e}}
\begin{document}

\title{Zero-Shot Distracted Driver Detection via \\Vision Language Models with Double Decoupling}

\author{
\IEEEauthorblockN{Takamichi Miyata}
\IEEEauthorblockA{
\textit{Chiba Institute of Technology}\\
Chiba, Japan \\
takamichi.miyata@it-chiba.ac.jp}
\and
\IEEEauthorblockN{Sumiko Miyata}
\IEEEauthorblockA{
\textit{Institute of Science Tokyo}\\
Tokyo, Japan \\
sumiko@ict.eng.isct.ac.jp}
\and
\IEEEauthorblockN{Andrew Morris}
\IEEEauthorblockA{
\textit{Loughborough University}\\
Loughborough, UK \\
A.P.Morris@lboro.ac.uk}
}

\maketitle

\begin{abstract}
Distracted driving is a major cause of traffic collisions, calling for robust and scalable detection methods. Vision-language models (VLMs) enable strong zero-shot image classification, but existing VLM-based distracted driver detectors often underperform in real-world conditions.
We identify subject-specific appearance variations (e.g., clothing, age, and gender) as a key bottleneck: VLMs entangle these factors with behavior cues, leading to decisions driven by who the driver is rather than what the driver is doing. 
To address this, we propose a subject decoupling framework that extracts a driver appearance embedding and removes its influence from the image embedding prior to zero-shot classification, thereby emphasizing distraction-relevant evidence.
We further orthogonalize text embeddings via metric projection onto Stiefel manifold to improve separability while staying close to the original semantics.
Experiments demonstrate consistent gains over prior baselines, indicating the promise of our approach for practical road-safety applications.
Code is available at https://github.com/mtakamichi/ZVL-DDD.

\end{abstract}


\section{Introduction}

Road traffic collisions remain a critical global safety issue. 
According to the World Health Organization, 1.35 million people die each year from road traffic collisions worldwide, and driver inattention is a major contributing factor~\cite{WHO18}. 
A study of 1,005 traffic collisions across six European countries reports that 32\% of cases involved drivers, passengers, or pedestrians judged to be distracted or inattentive~\cite{Talbot13}. 
These findings motivate distracted driving detection (DDD) as an essential component of driver monitoring, where early identification of risky behaviors and timely warnings can help mitigate traffic collisions.

Despite substantial progress, classic supervised learning approaches to DDD~\cite{Abouelnaga17, Saad20} remain impractical, as they require large-scale labeled in-cabin images that are costly to collect and annotate.
Moreover, these models are tightly coupled to the training distribution and often degrade sharply under dataset shift, including changes in camera viewpoint, illumination, and driver appearance. 
To address these limitations, recent work has explored the use of VLMs pretrained on massive image text pairs for zero-shot DDD~\cite{Radford21-CLIP}, namely performing inference without any task specific training on DDD.
However, even with strong pretrained VLMs, prior studies report that performance on public DDD benchmarks is still far from satisfactory~\cite{Hasan-22-DriveCLIP,Greer-24-DriverAC}, highlighting a persistent gap between the promise of VLMs and their practical accuracy for DDD.

We identify two key challenges in enabling zero-shot DDD with VLMs. First, existing VLMs are pretrained to be sensitive to visually salient differences. 
As general purpose models, when applied to DDD imagery they often respond strongly to subject specific factors such as driver identity, gender, age, and clothing. 
While such cues can be beneficial in applications such as person recognition and retrieval, they are detrimental for DDD, where the goal is to distinguish subtle behavior differences across drivers whose appearances can vary drastically. 
Second, as also noted in ReCLIP, fine-grained classification tasks like DDD typically rely on prompts with closely related semantics, which leads to clustered text embeddings and consequently degrades classification performance.
Figure~\ref{fig:tsne} previews this issue: naive zero-shot tends to organize embeddings by driver appearance and yields clustered text embeddings, while our decoupling reshapes the geometry toward class-centered structure.

To address these issues, we propose a framework that explicitly decouples driver appearance from behavior evidence and improves textual separability. 
Specifically, to reduce the influence of driver specific appearance, we modify each image embedding using the mean embedding vector computed for each driver, thereby suppressing identity related components and allowing the VLM to focus on distraction relevant cues. 
In addition, we transform class text embeddings into vectors that remain as close as possible to the original embeddings while becoming mutually orthogonal. 
This increases the separability of text representations and yields more reliable zero-shot classification for DDD.
The main contributions of this paper are summarized as follows:
\begin{itemize}
\item We discover that subject-specific appearance variations significantly hinder zero-shot distracted driver detection with vision-language models.
\item Based on the key insight, we propose a zero-shot DDD framework that decouples driver appearance from behavior evidence and orthogonalizes text embeddings to enhance separability.
\item We demonstrate consistent performance gains over prior VLM-based baselines on public distracted driver detection benchmarks, highlighting the promise of our approach for practical road-safety applications.
\end{itemize}

\section{Related Work}
\subsection{Vision-Language Models}

 Vision-Language Models (VLMs) learn aligned representations of images and natural language from diverse, large-scale image-text data, enabling open-vocabulary zero-shot classification beyond a fixed set of predefined classes. 
A representative line of work adopts a dual-encoder architecture, where an image encoder and a text encoder map their inputs into a shared embedding space and are trained on large-scale image-text pairs to maximize the similarity of matched image-text pairs while minimizing that of mismatched pairs~\cite{Radford21-CLIP,Zhai-23-SigLIP}. 
Through such contrastive pretraining, VLMs acquire transferable semantic knowledge that can be reused across downstream tasks with minimal or no task-specific supervision.
A common zero-shot classification scheme constructs a set of textual class descriptions, often referred to as text prompts, encodes them into text embedding vectors, and then predicts the class whose text embedding is most similar to the image embedding vector. 
However, the accuracy of zero-shot classification often degrades substantially on fine-grained recognition benchmarks, where categories differ only by subtle visual cues~\cite{Radford21-CLIP}.

\subsection{VLMs for Distracted Driver Detection}
 Hasan et al.\ conducted one of the earliest studies that systematically applied VLMs to DDD, evaluating both zero-shot prompting and supervised classifiers built on top of VLM image embeddings. They referred to this direction as DriveCLIP \cite{Hasan-22-DriveCLIP}. 
In their follow-up work~\cite{Hasan-24-VLM}, they further proposed methods that integrate information from multiple image frames to improve the performance of DriveCLIP. However, these results consistently indicate that naive zero-shot classification does not achieve satisfactory accuracy. Based on this observation, existing approaches have largely shifted toward full-shot learning, which can lead to overfitting to the target dataset and reduced generalization.

\subsection{Improving Zero Shot Capability of VLMs}
Recent work improves zero shot recognition by reshaping the embedding space of a pretrained VLM, rather than changing the backbone.
This is particularly relevant to fine grained settings, where semantically close prompts can yield poorly separated text embeddings and limit discrimination.
ReCLIP~\cite{Hu-24-ReCLIP} refines CLIP for a target domain in a source free manner using unlabeled target images and target class names.
It introduces a projection space to remove redundant dimensions and class agnostic information and to improve cross modal alignment, and then exploits neighborhood structure and cross modality self training to further refine the model.
PRISM~\cite{Molahasani-25-PRISM} targets spurious implicit biases inherited from large scale pretraining.
It uses an LLM to surface potential spurious correlations and learns an embedding projection that suppresses such correlations while preserving image text alignment, highlighting the effectiveness of lightweight projection based transformations for improving robustness.

\begin{figure*}[!t]
\centering
\includegraphics[width=0.9\textwidth]{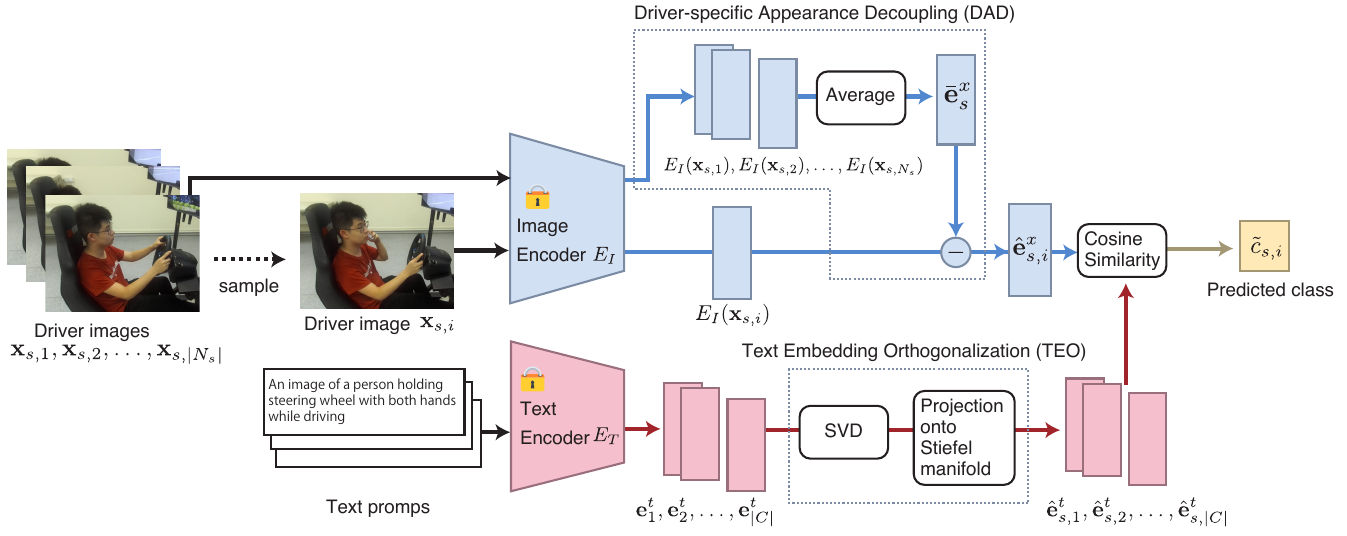}
\caption{Overview of our proposed double decoupling framework for zero-shot distracted driver detection. Given an input image captured by an in-cabin camera.}
\label{fig:pipeline}
\vspace{-0.5em}
\end{figure*}

\section{Method}
Given an in-cabin image $\mathbf{x}$, DDD aims to predict a class $c \in C$, where $C$ includes safe driving and distracted behaviors such as ``using a mobile phone''.
Following DriveCLIP~\cite{Hasan-22-DriveCLIP}, naive zero-shot classification predicts
\begin{equation}
\tilde{c}=\arg\max_{c \in C} \cos(E_I(\mathbf{x}), E_T(\mathbf{t}_c)),
\end{equation}
where $E_I$ and $E_T$ denote the image and text encoders of the VLM, respectively, $\mathbf{t}_c$ is the tokenized prompt for class $c$, and $\cos(\cdot, \cdot)$ is the cosine similarity.

Naive zero-shot DDD suffers from two major limitations: driver-specific appearance variations in image embeddings and poor separability among semantically close text embeddings.
Our method addresses these limitations through double decoupling, consisting of driver-specific appearance decoupling (DAD) in the image modality and text embedding orthogonalization (TEO) in the text modality, complemented by VLM-aware prompt engineering.
The overall pipeline is illustrated in Figure~\ref{fig:pipeline}.

\subsection{Driver-specific Appearance Decoupling (DAD)}
VLMs are known to respond sensitively to driver-specific appearance variations, which can degrade the accuracy of zero-shot DDD.
To mitigate this, we propose a method that modifies each image embedding using the mean embedding vector computed for each driver.
Specifically, given a set of images for driver $s\in S$, denoted as $\mathcal{X}_s=\{\mathbf{x}_{s,1}, \mathbf{x}_{s,2}, \ldots, \mathbf{x}_{s,N_s}\}$, we compute the mean image embedding vector for driver $s$, denoted as $\bar{\mathbf{e}}_s^x$, as follows:
\begin{equation}
\bar{\mathbf{e}}_s^x = \frac{1}{N_s} \sum_{i=1}^{N_s} E_I(\mathbf{x}_{s,i}),
\end{equation}
where, $N_s$ is the number of images for driver $s$.
Next, we modify the embedding vector of each image $\mathbf{x}_{s,i}$ as follows:
\begin{equation}
\hat{\mathbf{e}}_{s,i}^x = E_I(\mathbf{x}_{s,i}) - \bar{\mathbf{e}}_s^x,
\end{equation}
The modified embedding vector $\hat{\mathbf{e}}_{s,i}^x$ is then used in the same manner as in naive zero-shot classification.

\subsection{Text Embedding Orthogonalization (TEO)}
ReCLIP~\cite{Hu-24-ReCLIP} has shown that refining text embeddings to enhance inter-class separability can improve zero-shot classification performance.
Inspired by this, we propose a method to orthogonalize text embedding vectors using metric projection onto the Stiefel manifold.
Specifically, given a set of text embedding vectors $\{\mathbf{e}_c^t = E_T(\mathbf{t}_c) | c \in C\}$, we seek a set of orthogonal vectors $\{\hat{\mathbf{e}}_c^t | c \in C\}$ that minimize the following objective:
\begin{equation}
\min_{\{\hat{\mathbf{e}}_c^t\}} \sum_{c \in C} \|\hat{\mathbf{e}}_c^t - \mathbf{e}_c^t\|^2 \quad \text{s.t.} \quad \hat{\mathbf{e}}_c^{t\top} \hat{\mathbf{e}}_{c'}^t = 0 \quad \forall c \neq c'.
\end{equation}
This optimization problem can be efficiently solved using SVD of the matrix $T$, which is formed by stacking the original text embeddings:
\begin{equation}
T = [\mathbf{e}_{1}^t, \mathbf{e}_{2}^t, \ldots, \mathbf{e}_{{|C|}}^t].
\end{equation}
Performing singular value decomposition (SVD) on $T$ yields $T = U \Sigma V^\top$, where $U$ and $V$ are orthogonal matrices and $\Sigma$ is a diagonal matrix of singular values. The orthogonalized text embeddings are then given by the columns of $U V^\top$.  
The resulting orthogonal text embeddings $\{\hat{\mathbf{e}}_c^t | c \in C\}$. $\hat{\e}_{s,i}^x$ and $\hat{\e}_c^t$  are then used for zero-shot classification for $i$-th image of driver subject $s$, i.e.,
\begin{equation}
\tilde{c}_{s,i} = \arg \max_{c \in C} \cos(\hat{\e}_{s,i}^x, \hat{\e}_c^t).
\label{eq:final_prediction}
\end{equation}

\subsection{Prompt Engineering for DDD based on VLM characteristics}
\label{sec:prompt_engineering}

To improve zero-shot DDD, we design class prompts based on the visual and linguistic characteristics of VLMs.
Since zero-shot predictions can be sensitive to prompt wording, we follow three principles.
First, we make the defining visual evidence explicit, especially for the non-distracted class (P-1).
Second, we specify laterality only when it provides discriminative visual evidence (P-2).
For example, in phone-related actions, we explicitly describe the less common and visually diagnostic left-hand case, while omitting ``right'' when it is not essential, as this can reduce prompt-level bias toward laterality.
Third, we use standard, visually grounded action descriptions and avoid ambiguous or uncommon expressions (P-3).
Dataset-specific class descriptions and their correspondence to these principles are given in the experimental section.

\begin{figure*}[!t]
\centering
\subfloat[Naive zero-shot (DriveCLIP).]{
  \includegraphics[width=0.35\textwidth]{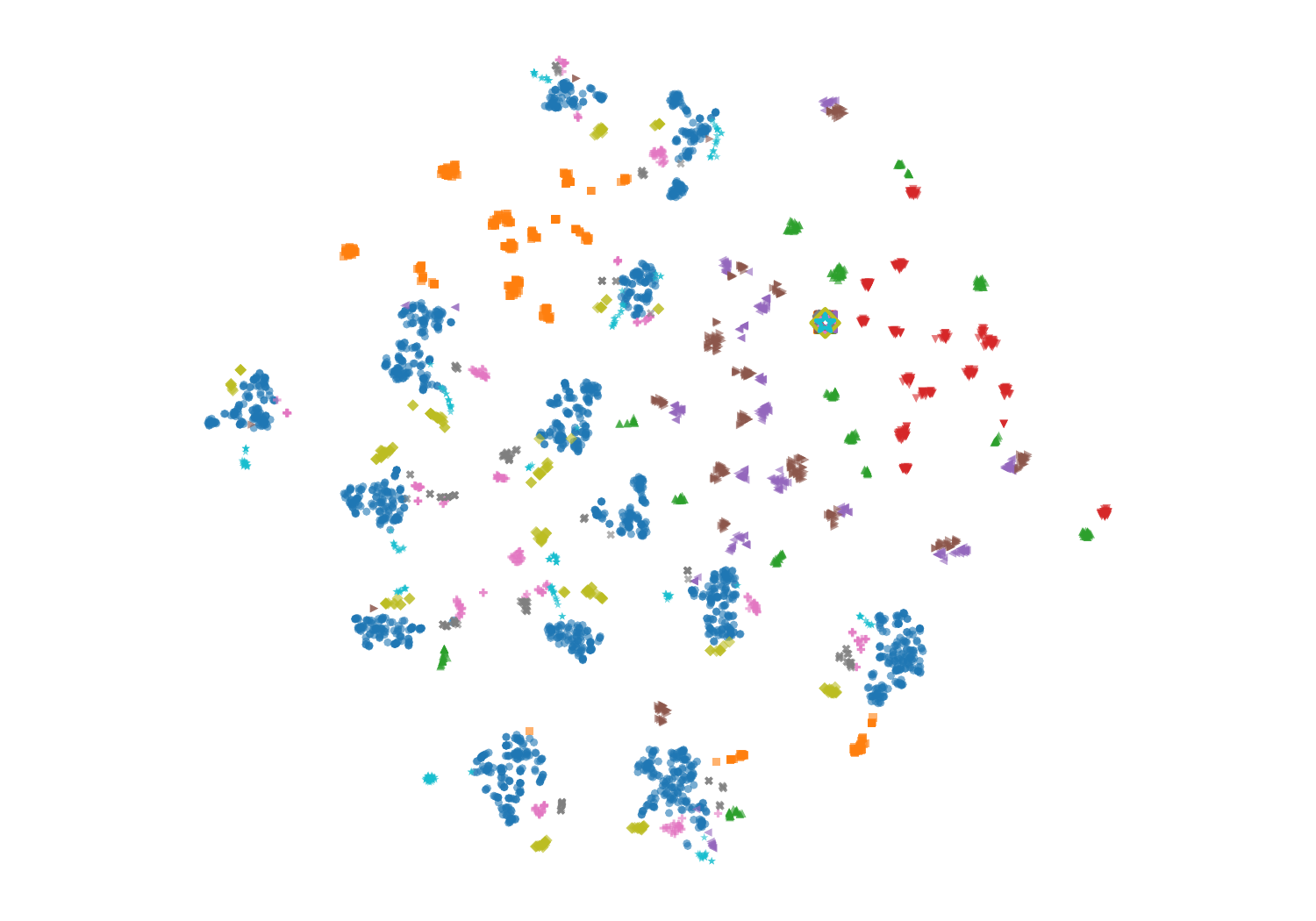}
  \label{fig:tsne:a}
}
\hfill
\subfloat[Ours]{
  \includegraphics[width=0.5\textwidth]{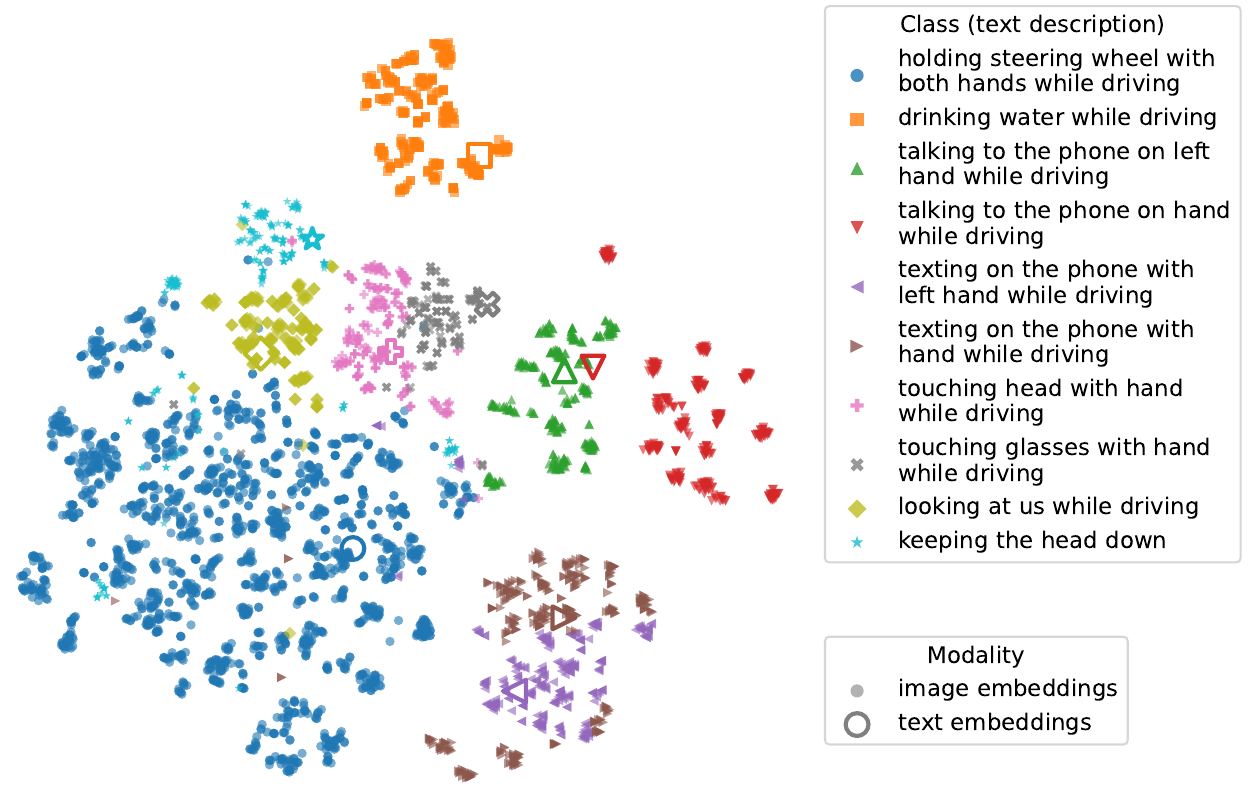}
  \label{fig:tsne:b}
}
\caption{t-SNE visualization of joint embedding spaces on SAM-DD. We randomly sample 30\% of the test images and embed them with the CLIP image encoder, and we also embed all class prompts with the CLIP text encoder. Colors and marker shapes indicate classes, and hollow markers indicate text embeddings while filled markers indicate image embeddings. (a) Naive zero-shot exhibits strong clustering by driver appearance rather than by behavior class, and the text embeddings collapse into a narrow region. (b) With DAD and TEO, appearance-driven dispersion is reduced and class-wise clusters become more pronounced, while text embeddings spread out and move closer to the corresponding image clusters.}
\label{fig:tsne}
\vspace{-0.5em}
\end{figure*}

\section{Experiment and Results}

{\bf Baselines.}
We compare our proposed method against the naive zero-shot DDD of DriveCLIP~\cite{Hasan-22-DriveCLIP,Hasan-24-VLM}.
Hereafter, we refer it simply as DriveCLIP.

{\bf Datasets.}
We conduct experiments on
Singapore AutoMan@NTU distracted driving (SAM-DD) dataset~\cite{Yang-24-SAM-DD} and the State Farm distracted driver detection (StateFarm) dataset~\cite{StateFarm}.
Both datasets contain in-cabin driver images captured under diverse conditions and provide annotations for 10 driving-behavior classes, including safe driving and various distracted actions.

{\bf Metrics.} 
For 10-class Classification, we evaluate the performance using standard classification metrics including Top-1 accuracy, Top-3 accuracy, precision, and recall.
In practical applications, it is also important to evaluate binary classification accuracy between the safe driving class ($c=0$) and distracted classes ($c\in \{1,\ldots,9\}$).
The binary prediction is made by predicting safe driving if the class with the highest similarity in Eq.~\eqref{eq:final_prediction} is the safe driving class, and distracted otherwise.
To evaluate the accuracy of the binary classification, we compute the area under the precision-recall curve (AUPRC), which summarizes precision and recall across different thresholds. 
Moreover, we also report the false negative rate (FNR) to indicate the rate of missed distracted detections.

{\bf Implementation Details.}
We use CLIP~\cite{Radford21-CLIP} with ViT-L/14@336px backbone as the VLM for all experiments.
For image preprocessing, input images are resized to $336 \times 336$ pixels and normalized according to CLIP's specifications.
We implement our method using PyTorch and run experiments on a single NVIDIA RTX 4090 GPU.

{\bf Prompt Engineering.}
For SAM-DD, the prompt modifications correspond to the three principles (P-1, P-2, and P-3) as follows.
Following P-1, we replace \texttt{driving safely} with the explicit visual cue \texttt{holding steering wheel with both hands while driving}.
Following P-2, we remove unnecessary right-hand laterality from phone-related classes while preserving the behavior description, since explicitly saying ``right'' may overemphasize laterality rather than the target action.
Following P-3, we rewrite ambiguous or uncommon expressions into more standard or visually grounded descriptions, including \texttt{touching hairs} to \texttt{touching head}, \texttt{adjusting glasses} to \texttt{touching glasses}, \texttt{reaching behind} to \texttt{looking at us while driving}, and \texttt{dropping the head} to \texttt{keeping the head down}.
For State Farm, prompts are designed using the same principles.
The full prompt list will be included in the released implementation.
These refinements are complementary to our decoupling strategies and are applied consistently across all experiments.

\begin{table}[t]
\centering
\caption{Comparison of zero-shot prompt class names for DDD. 
This corresponds to the 10 classes of the SAM-DD dataset~\cite{Yang-24-SAM-DD}.
 Differences in class names are highlighted in bold on the right column.
}
\label{tab:prompt_comparison}
\begin{tabularx}{\linewidth}{X X}
\toprule
DriveCLIP class name \cite{Hasan-22-DriveCLIP} & Ours class name \\
\midrule
\texttt{driving safely} & \textbf{\texttt{holding steering wheel with both hands while driving}} \\ \arrayrulecolor{gray!40}\hline
\texttt{drinking water while driving} & \texttt{drinking water while driving} \\ \arrayrulecolor{gray!40}\hline
\texttt{talking to the phone on left hand while driving} & \texttt{talking to the phone on left hand while driving} \\ \arrayrulecolor{gray!40}\hline
\texttt{talking to the phone on right hand while driving} & \textbf{\texttt{talking to the phone on hand while driving}} \\ \arrayrulecolor{gray!40}\hline
\texttt{texting on the phone with left hand while driving} & \texttt{texting on the phone with left hand while driving} \\ \arrayrulecolor{gray!40}\hline
\texttt{texting on the phone with right hand while driving} & \textbf{\texttt{texting on the phone with hand while driving}} \\ \arrayrulecolor{gray!40}\hline
\texttt{touching hairs with hand while driving} & \textbf{\texttt{touching head with hand while driving}} \\ \arrayrulecolor{gray!40}\hline
\texttt{adjusting glasses with hand while driving} & \textbf{\texttt{touching glasses with hand while driving}} \\ \arrayrulecolor{gray!40}\hline
\texttt{reaching behind while driving} & \textbf{\texttt{looking at us while driving}} \\ \arrayrulecolor{gray!40}\hline
\texttt{dropping the head while driving} & \textbf{\texttt{keeping the head down}} \\ \arrayrulecolor{black}
\bottomrule
\end{tabularx}
\end{table}

\subsection{Main Results}

Table~\ref{tab:ddd_main_results} presents the performance comparison between DriveCLIP and our proposed method on the SAM-DD and State Farm datasets.
Our method achieves consistent improvements across all evaluation metrics on both datasets, demonstrating the effectiveness of the proposed decoupling strategies and prompt engineering.
Notably, on SAM-DD, we observe a substantial increase in Top-1 accuracy from 66.5\% to 75.9\%, indicating that our approach enables more reliable zero-shot distracted driver detection.
These results highlight the potential of our method for practical applications in enhancing road safety through improved driver monitoring systems.

Table~\ref{tab:classwise_performance} shows the class-wise performance of our method on the SAM-DD dataset.
We observe that classes involving visually distinctive objects, such as drinking water and phone-related actions, generally perform better, indicating that the model can effectively use object cues for classification.
Conversely, classes that rely more heavily on body or head pose recognition, such as reaching behind or dropping the head, exhibit lower recall and precision. This suggests that these behaviors are more difficult for zero-shot CLIP to detect accurately.

Examining the confusion matrices for each driver separately revealed that baseline errors were strongly dependent on both the target class and the driver identity. For Val13 in SAM-DD, DriveCLIP misclassified several distracted behaviors as safe driving. Specifically, in the categories of texting with the right hand, touching one's hair, and reaching behind, 93.3\% of the images were predicted as safe driving. In contrast, the proposed method achieved a micro-accuracy of 94.4\% for the same three classes. These results demonstrate that the DriveCLIP decision is influenced by driver-specific appearance and imaging conditions rather than by driving behavior itself.

\begin{table}[t]
\centering
\caption{Performance comparison on two datasets.}
\label{tab:ddd_main_results}
\begin{tabular}{llcccc}
\toprule
Method & Dataset &Top-1$\uparrow$ & Top-3$\uparrow$ & Rec.$\uparrow$ & Prec.$\uparrow$ \\
\midrule
DriveCLIP~\cite{Hasan-24-VLM} & SAM-DD &66.5 & 85.8 & 44.8 & 44.7 \\
\textbf{Ours} & SAM-DD & \textbf{75.9}  &\textbf{96.9} & \textbf{68.4} & \textbf{70.4} \\ \arrayrulecolor{gray!40} \midrule
DriveCLIP~\cite{Hasan-24-VLM} & Statefarm &45.5 & 76.6 &  44.4 &  48.4 \\
\textbf{Ours} & Statefarm & \textbf{54.6}  &\textbf{89.3} & \textbf{54.6} & \textbf{55.7} \\ \arrayrulecolor{black}
\bottomrule
\end{tabular}
\end{table}

\begin{table}[t]
\centering
\caption{Class-wise performance on the SAM-DD dataset.}
\label{tab:classwise_performance}
\begin{tabular}{lcc}
\hline
Class & Rec.$\uparrow$  & Prec.$\uparrow$  \\
\hline
driving safely & 0.864 & 0.864 \\
drinking water & 0.998 & 0.980 \\
talking - left & 0.432 & 0.989 \\
talking - right & 0.999 & 0.657 \\
texting - left  & 0.403 & 0.658 \\
texting - right & 0.755 & 0.613 \\
touching hairs & 0.703 & 0.760 \\
adjusting glasses & 0.863 & 0.616 \\
reaching behind & 0.513 & 0.633 \\
dropping the head & 0.307 & 0.269 \\
\hline
Macro avg. & 0.684 & 0.704 \\
\hline
\end{tabular}
\end{table}

\subsection{Performance on Binary Classification}
To further assess the effectiveness of our method in distinguishing between distracted and non-distracted driving, we conduct a binary classification evaluation by grouping all distracted driving classes into a single distracted category.
Table~\ref{tab:2c_results} summarizes the results of this evaluation.
Our method outperforms DriveCLIP significantly, achieving a 2C-AUPRC of 95.8 compared to 90.6 and a 2C-FNR of 10.9 compared to 32.6.
These findings underscore the robustness of our approach in accurately identifying distracted driving behaviors, which is critical for real-world driver monitoring applications.

\begin{table}[t]
\centering
\caption{Two-class evaluation results.}
\label{tab:2c_results}
\begin{tabular}{llcc}
\toprule
Method & Dataset & 2C-AURPC$\uparrow$ & 2C-FNR$\downarrow$ \\
\midrule
DriveCLIP~\cite{Hasan-24-VLM} & SAM-DD & 90.6 & 32.6 \\
\textbf{Ours} & SAM-DD & \textbf{95.8} & \textbf{10.9} \\ \arrayrulecolor{gray!40} \midrule
DriveCLIP~\cite{Hasan-24-VLM} & Statefarm & 95.6 & 20.9 \\
\textbf{Ours} & Statefarm & \textbf{97.1} & \textbf{11.9} \\ \arrayrulecolor{black}
\bottomrule
\end{tabular}
\end{table}

\subsection{Visualization of the Embedding Vectors}
To better understand why naive zero-shot DDD fails and how our double decoupling improves it, we visualize the embedding geometry using t-SNE with a perplexity of 30.
Figure~\ref{fig:tsne} shows a joint two-dimensional projection of image embeddings and text embeddings.
We compute image embeddings from a random 30\% subset of the SAM-DD test set, and we compute text embeddings from the prompts of all classes.
In the plot, classes are distinguished by color and marker shape, and the modality is indicated by whether markers are filled (image) or hollow (text).

In Fig.~\ref{fig:tsne}(a), although a few behavior classes form small clusters, the overall structure is dominated by subject-specific appearance variations, resulting in groupings that are more consistent with driver identity than with behavior class.
This supports our observation that VLMs can overreact to visually salient driver attributes, making fine-grained behavior discrimination difficult even with carefully chosen prompts.
Moreover, the text embeddings occupy a very narrow region and are highly clustered, which is consistent with the phenomenon reported in ReCLIP~\cite{Hu-24-ReCLIP} where text embeddings can collapse and lose inter-class separability.

In Fig.~\ref{fig:tsne}(b), DAD reduces dispersion caused by driver appearance, and the image embeddings become more class-centered, yielding clearer class-wise clusters.
At the same time, TEO spreads the text embeddings and moves them closer to the corresponding image clusters.
Together, these effects provide an intuitive explanation for the consistent performance gains of our method.

\subsection{Ablation Studies}

To demonstrate the effectiveness of each component of our proposed method, we conducted ablation studies.
As shown in Table~\ref{tab:ablation}, no single component consistently improves performance.
Conversely, partial combinations improve macro recall and precision, though they reduce Top-1 accuracy. 
This is partly due to the class imbalance in SAM-DD; losing correct predictions for the "safe driving" class can reduce micro-level accuracy.
For instance, in PE+DAD, subtracting the subject-wise mean image embedding alters the angular structure of the image embeddings by emphasizing subject-relative residual features.
However, without TEO, the text embeddings remain angularly close, which can increase class confusion and lead to errors in the safe-driving class.
The full PE+DAD+TEO configuration resolves this issue and achieves the best performance across all metrics.

\begin{table}[t]
\centering
\caption{Ablation study on DDD. Best results are in bold, and second-best results are underlined.}
\label{tab:ablation}
\begin{tabular}{lcccccc}
\toprule
Method & PE & DAD & TEO & Top-1 & Rec. & Prec. \\
\midrule
DriveCLIP &  &  &  & \underline{66.5} & 44.8 & 44.7 \\
- & \checkmark &  &  & 66.3 & 40.2 & 50.3 \\
- &  & \checkmark &  & 52.5 & 28.9 & 27.9 \\
- & &  & \checkmark  & 65.2 & 39.6 & 53.2 \\
- & \checkmark & \checkmark &  & 57.2 & 63.5 & 66.1 \\
- & \checkmark &  & \checkmark  & 56.5 & \underline{64.2} & \underline{66.9} \\
- &  & \checkmark & \checkmark  & 63.2 & 63.2 & 51.2 \\
\midrule    
{\bf Ours}  & \checkmark & \checkmark & \checkmark & \textbf{75.9} & \textbf{68.4} & \textbf{70.4} \\
\bottomrule
\end{tabular}
\end{table}

\subsection{Runtime Analysis}
We further evaluate the runtime feasibility of the proposed method for edge-oriented deployment.
Compared with DriveCLIP, our method introduces only lightweight post-processing operations, and its overhead is negligible because the CLIP image encoder dominates the overall inference cost.
On an NVIDIA RTX 4090 GPU, both DriveCLIP and our method require approximately 10 ms per frame.
MobileCLIP2-L/14 has also been reported to achieve 57.9 ms per frame on an iPhone 12 Pro Max~\cite{mobileclip2}, suggesting that CLIP-style VLMs can operate at around 15 Hz on mobile edge hardware.
These results support the practical feasibility of our CLIP-style approach for real-time edge deployment.

\section{Discussion}
Our results demonstrate that naive zero-shot DDD with pretrained VLMs is fundamentally limited by two forms of entanglement. Image embeddings are strongly shaped by subject-specific appearance, and text embeddings for fine-grained prompts often collapse into a narrow region, which weakens class separability. 
We address both issues with a lightweight double decoupling strategy that removes appearance-driven components from visual embeddings and increases the separability of textual embeddings. 
This approach yields consistent and substantial gains over prior VLM-based baselines in both multi-class and two-class evaluations. The t-SNE visualization supports this interpretation and offers an intuitive explanation for the improved zero-shot accuracy.

Recent work has explored zero-shot DDD with large generative VLMs that couple an LLM with a vision backbone~\cite{Greer-24-DriverAC,Hu-25-MultimodalLLM,Wang25-Leveraging}.
Although these models are promising, their computational and memory requirements can limit their practicality for onboard driver-monitoring systems.
Our preliminary test using Llama-3.2-11B~\cite{meta-llama-32-vision-11b} required approximately 3 s per frame on the RTX 4090, which is roughly 300 times slower than our proposed method.
In contrast, as shown in our runtime analysis, the proposed method adds negligible overhead to DriveCLIP while retaining the efficiency of CLIP-style zero-shot inference.
These characteristics make our approach better aligned with  edge-deployment constraints.

DAD requires a driver-specific mean embedding, which can be estimated from unlabeled frames during the initialization period. Since no behavior labels are required, this calibration can be performed automatically. However, the required number of frames, robustness to appearance changes, and online updating are still to be determined.

Although our method significantly improves zero-shot performance, relying solely on camera images to issue driver warnings can still be risky given the current level of reliability. 
Therefore, in practical deployments, the output of our proposed method should be treated as one component of a broader decision system, combined with additional contextual signals, such as vehicle speed, acceleration, and steering behavior.

\section{Conclusion}
We presented a zero-shot framework for distracted driving detection using vision-language models, built around double decoupling in the visual and textual embedding spaces. By suppressing driver-appearance factors in image embeddings and improving separability of class text embeddings, our approach consistently improves zero-shot performance over prior VLM-based baselines on SAM-DD, while remaining lightweight and training-free on the target task. In future work, we will replace the backbone with more recent VLMs such as SigLIP 2~\cite{Zhai-23-SigLIP} and conduct a more comprehensive evaluation across additional datasets and settings to further assess robustness under broader domain shifts.

\section*{Acknowledgment}
This work was supported by JSPS KAKENHI Grant Number 23K03871, research grant from the Telecommunications Advancement Foundation, and travel grant from the Kajima Foundation.

\input{CSNDSP_2026_ZVLDDD.bbl}

\end{document}

%% file: CSNDSP_2026_ZVLDDD.bbl
% Generated by IEEEtran.bst, version: 1.14 (2015/08/26)